\documentclass[moor]{informs2}

\long\def\hide#1{}

\usepackage{subfigure}
\usepackage{comment}
\usepackage{algorithm}
\usepackage{algpseudocode}
\usepackage{enumerate}
\usepackage{type1cm}
\usepackage[colorlinks=true,breaklinks=true,bookmarks=true,urlcolor=blue,citecolor=blue,linkcolor=blue,bookmarksopen=false,draft=false]{hyperref}
\usepackage[usenames,dvipsnames,svgnames,table]{xcolor}

\newcommand{\eProof}{\hspace*{.1in} \hfill
\begin{picture}(6,6)
\thicklines \put(0,0){\line(0,7){7}} \put(1,0){\line(0,7){7}}
\put(1.5,0){\line(0,7){7}} \put(2,0){\line(0,7){7}}
\put(3,0){\line(0,7){7}} \put(4.5,0){\line(0,7){7}}
\put(4,0){\line(0,7){7}} \put(5,0){\line(0,7){7}}
\end{picture} }

\usepackage{natbib}
 \NatBibNumeric
 \def\bibfont{\small}%
 \def\bibsep{\smallskipamount}%
 \def\bibhang{24pt}%
 \bibpunct[, ]{[}{]}{,}{n}{}{,}%

\newboolean{showcomments}
\setboolean{showcomments}{true}
\newcommand{\raga}[1]{  \ifthenelse{\boolean{showcomments}}
{ \textcolor{Red}{(Raga says:  #1)}} {}  }
\newcommand{\partha}[1]{\ifthenelse{\boolean{showcomments}}
{ \textcolor{Blue}{(Partha says: #1)} } {} }
\newcommand{\abhishek}[1]{\ifthenelse{\boolean{showcomments}}
{ \textcolor{Green}{(Abhishek says: #1)} } {} }
\newcommand{\arpita}[1]{\ifthenelse{\boolean{showcomments}}
{ \textcolor{Magenta}{(Arpita says: #1)} } {} }

\makeatletter
\newcommand{\subalign}[1]{%
  \vcenter{%
    \Let@ \restore@math@cr \default@tag
    \baselineskip\fontdimen10 \scriptfont\tw@
    \advance\baselineskip\fontdimen12 \scriptfont\tw@
    \lineskip\thr@@\fontdimen8 \scriptfont\thr@@
    \lineskiplimit\lineskip
    \ialign{\hfil$\m@th\scriptstyle##$&$\m@th\scriptstyle{}##$\crcr
      #1\crcr
    }%
  }
}
\makeatother

\def\EMAIL#1{\href{mailto:#1}{#1}}

\TheoremsNumberedThrough
\ECRepeatTheorems

\EquationsNumberedThrough

\MANUSCRIPTNO{XXX}

\begin{document}
%%%%%%%%%%%%%%%%

\RUNAUTHOR{Biswas, Gopalakrishnan, and Dutta}

\RUNTITLE{Managing Overstaying Electric Vehicles in Park-and-Charge Facilities}

\TITLE{Managing Overstaying Electric Vehicles in Park-and-Charge Facilities}

\ARTICLEAUTHORS{
\AUTHOR{Arpita Biswas}
\AFF{Xerox Research Centre India, Bangalore, Karnataka 560103,\\ \EMAIL{Arpita.Biswas@xerox.com}}
\AUTHOR{Ragavendran Gopalakrishnan}
\AFF{Xerox Research Centre India, Bangalore, Karnataka 560103,\\ \EMAIL{Ragavendran.Gopalakrishnan@xerox.com}}
\AUTHOR{Partha Dutta}
\AFF{Xerox Research Centre India, Bangalore, Karnataka 560103,\\ \EMAIL{Partha.Dutta@xerox.com}}
}

\ABSTRACT{
With the increase in adoption of Electric Vehicles (EVs), proper utilization of the charging infrastructure is an emerging challenge for service providers. Overstaying of an EV after a charging event is a key contributor to low utilization. Since overstaying is easily detectable by monitoring the power drawn from the charger, managing this problem primarily involves designing an appropriate ``penalty'' during the overstaying period. Higher penalties do discourage overstaying; however, due to uncertainty in parking duration, less people would find such penalties acceptable, leading to decreased utilization (and revenue). To analyze this central trade-off, we develop a novel framework that integrates models for realistic user behavior into queueing dynamics to locate the optimal penalty from the points of view of utilization and revenue, for different values of the external charging demand. Next, when the model parameters are unknown, we show how an online learning algorithm, such as UCB, can be adapted to learn the optimal penalty. Our experimental validation, based on charging data from London, shows that an appropriate penalty can increase both utilization and revenue while significantly reducing overstaying.
}

\KEYWORDS{park-and-charge; revenue optimization; sustainable behavior; multi-armed bandits}
\SUBJECTCLASS{Primary: Queues: applications, optimization; secondary: Probability: stochastic model applications}

\maketitle

\vspace{-0.125in}
\section{Introduction.}

As the number of on-road Electric Vehicles (EVs) increases rapidly, lack of adequate charging infrastructure is an area of growing concern. For example, a recent New York Times article reports that in California, scarcity of park-and-charge spots is a chronic problem: ``Electric-vehicle owners are unplugging one another's cars, trading insults, and creating black markets and side deals to trade spots in corporate parking lots'' \cite{NYTimesArticle}. While investing in a wide-spread deployment of charging stations by both public and private service providers is needed to address the infrastructure problem in the long-term, effective use of the existing charging infrastructure can help in significantly reducing this problem. In particular, curbing improper utilization of park-and-charge spots can improve their availability. Two prominent causes of utilization degradation are (i)~the overstaying problem, where an EV continues to occupy a park-and-charge spot even after it is fully charged, and (ii)~the ``icing'' problem, where a gas-powered car (Internal Combustion Engine or ICE vehicle) occupies a park-and-charge spot.

While icing and overstaying in park-and-charge spots are illegal in increasingly many jurisdictions, there is little or no enforcement \cite{NoEnforcementIcing}, as a result of which the frustrated users resort to ad-hoc measures ranging from mild (e.g., leaving courtesy notices) \cite{CourtesyNotice1,CourtesyNotice2} to drastic (e.g., publicly shaming the violator by posting a picture of the violation showing the violator's license plate on blogs and social media) \cite{PublicShaming1,PublicShaming2}. Extension cables that can help charge a vehicle parked a few spots away from an occupied park-and-charge spot are available in the market, but are very expensive \cite{ExtensionCable}. A longer term solution to these problems, however, will require both enforcement and an appropriate penalty for these events.

While enforcement with heavy penalty may help curb icing, a gentler approach is prudent to manage overstaying EVs. Depending on the demand for charging, overstaying EVs potentially block access to other EVs that might need charging, and so, it is important to discourage such behavior by imposing penalties. But, imposing too high a penalty might turn away EVs from using the park-and-charge facility altogether due to increased risk of a steep fine, since EV users may not exactly know their parking duration beforehand.

Our central contribution in this paper is a novel framework that combines a realistic user behavior model with traditional queueing dynamics to capture this trade-off and study the optimal penalty from the points of view of both utilization and revenue. Next, when the model parameters are unknown, we show how the well known UCB learning algorithm can be adapted to learn the optimal penalty over a period of time. Our experiments, based on charging data from London, show that an appropriate penalty results in increased utilization and significantly increased revenue. Also, perhaps surprisingly, we observe that the utilization achieved by imposing the revenue-maximizing penalty is very close to the maximum utilization.

\subsection{Related work.}

Dynamic pricing of parking has been an area of active study in the transportation literature. In~\cite{onnokdd,rowe2011}, the authors present dynamic pricing schemes for regular parking based on estimated demand to reduce both congestion and underuse. On the other hand, significant work has also been done on dynamic pricing of electric vehicle charging. In~\cite{tushar2012}, the authors model the problem as a Stackelberg game between the smart grid as the leader setting the price, and the vehicle owner as the follower deciding their charging strategies. In~\cite{anglin2013,gadh2012,hafner2009}, dynamic pricing of electricity for charging EVs is proposed based on the usage data in a given location and time. None of the above work, however, consider the problem of overstaying EVs. To the best of our knowledge, this is the first paper to investigate designing penalty schemes for park-and-charge facilities to combat overstaying EVs.

To address the situation where the probability distributions required for modelling user behaviour remain unavailable to the system until the users are actually observed, and the user behaviour needs to be learned over time, we model our problem as a Multi-Armed Bandit (MAB) problem, which has been well studied \cite{bubeck2012regret,auer2002finite,cesa2006prediction,agrawal1995sample} and applied to a wide variety of domains such as crowdsourcing, online advertising, dynamic pricing, and smart grids. However, we are not aware of any application of MABs to the park-and-charge scenario.

\section{Park-and-Charge system model.}

We adopt a simple three-part model for a parking area:
\begin{enumerate}[(a)]
\item Section~\ref{ssec:pricing-penalty} introduces the pricing function (during charging) and penalty function (during overstaying) for an EV.
\item Section~\ref{ssec:user-behavior} models how EV users respond to the posted penalty scheme, i.e., (i)~whether they agree to its terms and enter the parking area, and if so, (ii)~how long they stay.
\item Section~\ref{ssec:queueing} models the flow of EV users in and out of the parking area using queueing dynamics.
\end{enumerate}
We end the section by defining some performance measures of interest in Section~\ref{ssec:performance-metrics}.

\subsection{Pricing and penalty functions.}\label{ssec:pricing-penalty}

\begin{figure}[ht]
    \centering
    \includegraphics[width=0.4\columnwidth]{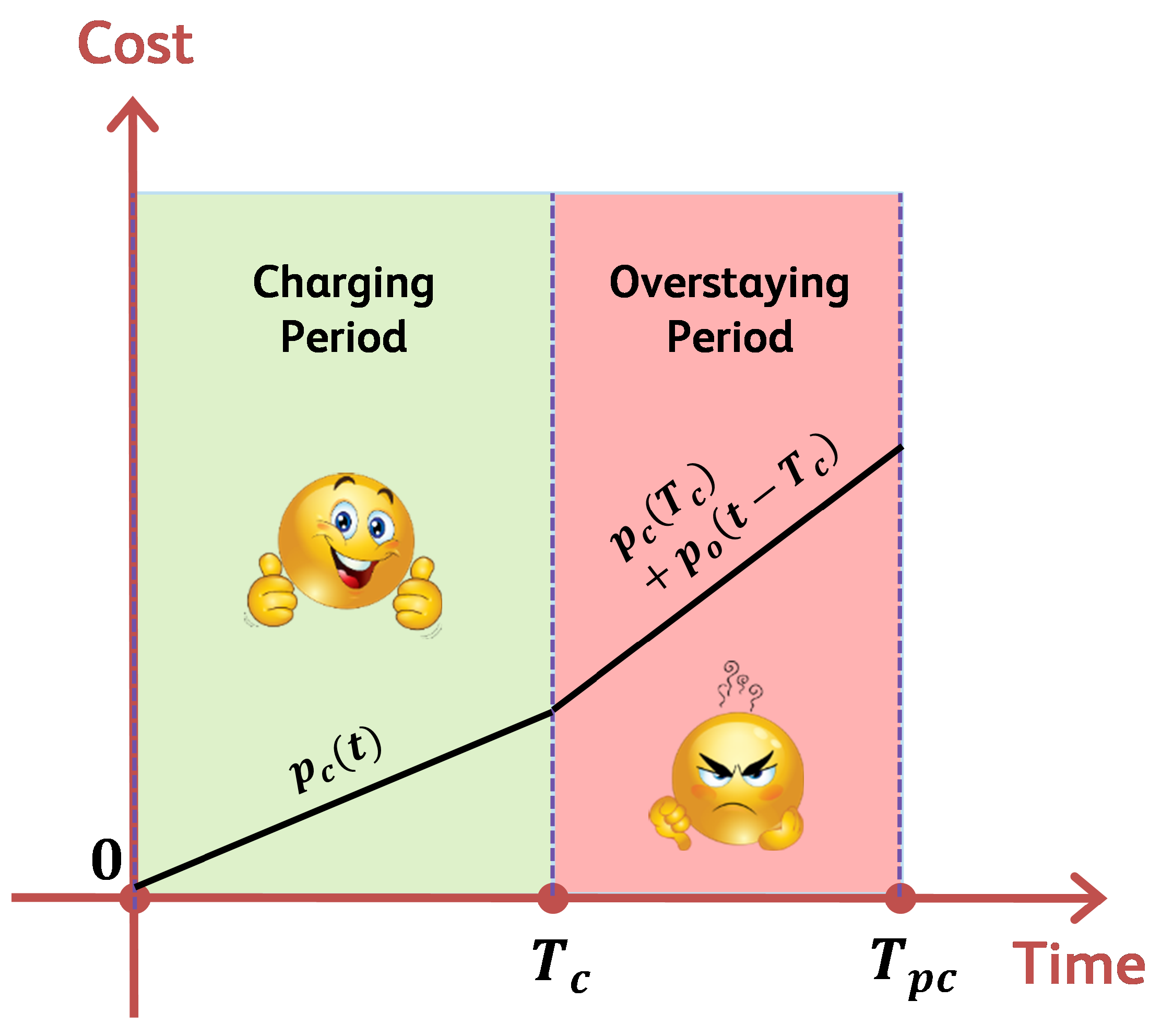}
    \caption{Pricing and penalty functions for an example scenario.\label{fig:price-penalty-functions}}
\end{figure}

Let the price function while actively charging be $p_{c}(t)$ and the penalty function for overstaying at a park-and-charge spot after charging is complete be $p_{o}(t)$, where $p_{c}(0)=p_{o}(0)=0$, and $p_{c}(t)$ and $p_{o}(t)$ are continuous, nondecreasing functions. In addition, we assume that $p_{o}(t)$ is a strictly increasing function, since we will need its inverse function, $p_{o}^{-1}$, to be well defined.\footnote{The framework can be extended to the case where $p_{o}(t)$ is not strictly increasing, by defining $p_{o}^{-1}(x) = \sup\{t:p_{o}(t)=x\}$. This could accommodate, for example, an initial ``grace period'' (no penalty) after charging is complete.} To be realistically implementable (easily understandable by an average user), these functions should not be more complicated than simple piecewise linear functions; however, our framework is general and can accommodate arbitrary functions. These functions are illustrated in Figure~\ref{fig:price-penalty-functions}.

\subsection{EV user behavior.}\label{ssec:user-behavior}

We define the following parameters, for each EV user:
\begin{itemize}
\item $T_c$, the duration of time it would take to fully charge their electric vehicle,
\item $T_a$, the length of their appointment, i.e., the EV user's preferred parking duration, and
\item $C_{\max}$, the ``penalty threshold'', i.e., the maximum penalty as a result of overstaying that they would bear for the convenience of an uninterrupted appointment.
\end{itemize}

Let $T_c$, $T_a$, $C_{\max}$ be independent, nonnegative random variables with finite means and cumulative distribution functions $F_c, F_a, F_{\max}$ respectively.\\

\noindent\textbf{Penalty acceptance probability:} When a user arrives at the parking area and inspects the posted penalty scheme, they know their $T_c$ and $C_{\max}$ values, but they may not know exactly how long their appointment will last; hence $T_a$ is not yet realized. However, the user can still estimate the likelihood that their penalty would not exceed $C_{\max}$, as follows:
\begin{equation}\label{eq:q}
q = Pr\left(p_{o}\left(T_a-T_c\right)\leq C_{\max}\right) = F_a(T_c + p_{o}^{-1}(C_{\max})).
\end{equation}
Thus, we assume that an arriving user will enter the parking lot with probability $q$. Let $\overline{q}$ denote its mean, given by:
\begin{equation}\label{eq:qbar}
\overline{q}=\int\int F_a\left(T_c+p_{o}^{-1}(C_{\max})\right)dF_c dF_{\max}.
\end{equation}

\noindent\textbf{Actual parking duration:} If a user decides to enter the parking area, they park their EV at a park-and-charge spot and leave for their appointment. $T_a$ is then realized, and if it exceeds $T_c + p_{o}^{-1}(C_{\max})$, we assume that the user reluctantly interrupts their appointment to avoid paying more than $C_{\max}$ penalty. Thus, the actual parking duration is given by:\footnote{If the realized value of $T_{a}$ is less than $T_c$, we assume that the user leaves immediately after $T_a$ time units, without waiting for their EV to finish charging.}
\begin{equation}\label{eq:tpc}
T_{pc} = \min\left\{T_c + p_{o}^{-1}(C_{\max}),T_a\right\},
\end{equation}
and hence, the duration of time the EV overstays is given by:
\begin{equation}\label{eq:to}
T_{o} = \left(T_{pc}-T_c\right)^+ = \max\{T_{pc}-T_c,0\}.
\end{equation}

\noindent\textbf{Revenue collected:} The revenue from an EV's time at the park-and-charge spot (the cost to the EV user) is given by:
\begin{equation}\label{eq:rev}
R = p_{c}(T_{pc}-T_{o}) + p_{o}(T_{o}).
\end{equation}

\subsection{Queueing model.}\label{ssec:queueing}

Since there is a finite number of park-and-charge spots in the parking area, the impact of overstaying depends on the arrival process to the parking lot, e.g., if the inter-arrival times are uniformly distributed with a low arrival rate, a limited amount of overstaying may be harmless, as opposed to a high arrival rate during weekends at a mall. We capture such phenomena using an $M$/$G$/$N$/$N$ queueing model~\cite{Kleinrock75}, where the EVs that accept the posted penalty scheme and enter the parking lot are the ``jobs'', and the $N$ park-and-charge spots are the ``servers''. The parameters and assumptions are as follows:
\begin{itemize}
\item EVs arrive at the parking area according to a Poisson process with rate $\lambda$. However, only those EVs whose users accept the posted penalty scheme (with probability $q$, given by~\eqref{eq:q}), and enter the parking area are counted for further analysis. Using the total probability theorem and the well known ``Poisson splitting'' property, it can be shown (see Proposition~\ref{prop:poisson} below) that this ``filtered'' arrival process is still Poisson, with rate $\lambda\ \overline{q}$, where $\overline{q}$ is given by~\eqref{eq:qbar}.
\item An arriving EV that finds all $N$ park-and-charge spots occupied leaves the system (there is no waiting). Otherwise, the EV stays at a park-and-charge spot for a duration of $T_{pc}$, given by~\eqref{eq:tpc}; thus, the service times are i.i.d. according to the cumulative distribution function of $T_{pc}$.\footnote{This could be a general distribution; however, in an $M$/$G$/$N$/$N$ system, the stationary distribution of its Markov chain (that tracks the number of EVs) depends on the service time distribution only through its mean.}
\end{itemize}

If $N_{pc}$ is the random variable denoting the number of EVs in the parking area, then its steady state distribution is given by:
\begin{equation}\label{eq:mgnn}
Prob\left(N_{pc}=i\right) = \frac{\rho^i/i!}{\sum_{j=0}^{N} \rho^j/j!},\quad 0\leq i\leq N,
\end{equation}
where $\rho=\lambda\ \overline{q}\ \mathbb{E}[T_{pc}]$ is the ``offered load''. Thus, in steady state, the mean number of EVs in the parking area is:
\begin{equation}\label{eq:meanNpc}
\mathbb{E}[N_{pc}] = \rho\left(1-\frac{\rho^N/N!}{\sum_{j=0}^{N} \rho^j/j!}\right).
\end{equation}

\begin{proposition}\label{prop:poisson}
Let $X(t)$ be a Poisson process with rate $\lambda$, and let each arrival $i$ counted by this process be associated with a parameter $Y_i$ that is drawn independently for each $i$ from a common distribution $Y$. If $X'(t)$ is a filtered process that counts each $X(t)$-arrival $i$ with probability $h(Y_i)$, then $X'(t)$ is also a Poisson process with rate $\lambda\ \overline{h}$, where $\overline{h}=\mathbb{E}_Y[h(Y)]$.
\end{proposition}

\proof{Proof.}
The probability with which a random $X(t)$-arrival is counted is given by
\begin{equation*}
\begin{split}
Prob\left(\mbox{arrival $i$ is counted}\right) &= \int Prob\left(\mbox{arrival $i$ is counted}\ |\ Y_i=y\right) dF_Y(y)\\
&= \int h(y)dF_Y(y) = \overline{h},
\end{split}
\end{equation*}
where $F_Y$ denotes the cumulative distribution function of $Y$. The first step follows from the total probability theorem, and the last step is the definition of expectation. It then follows from the well known Poisson splitting property that the filtered process $X'(t)$ is also Poisson with rate $\lambda\ \overline{h}$.
\eProof

\subsection{Performance measures.}\label{ssec:performance-metrics}

We define the following performance measures of interest. Throughout, $T>0$ denotes an arbitrary duration of time.
\begin{enumerate}[(a)]
\item \textbf{Throughput:} The throughput of the park-and-charge system, denoted by $\tau$, is the average rate at which EV users leave the park-and-charge area, given by:
    \begin{equation}\label{eq:throughput}
    \tau = \frac{\mathbb{E}[N_{pc}]}{\mathbb{E}[T_{pc}]}.
    \end{equation}
\item \textbf{Overstay:} The fraction of time spent overstaying at park-and-charge spots, denoted by $\gamma_o$, is given by:
    \begin{equation}\label{eq:overstaying}
    \gamma_o = \frac{\mbox{Overstaying Time}}{\mbox{Total Time}} = \frac{\tau\ T\ \mathbb{E}[T_o]}{N\ T} = \frac{\mathbb{E}[N_{pc}]}{N}\frac{\mathbb{E}[T_o]}{\mathbb{E}[T_{pc}]}.
    \end{equation}
\item \textbf{Utilization:} The utilization, denoted by $\gamma_u$, is defined as the fraction of time the park-and-charge spots were used for charging, and is given by:
    \begin{equation}\label{eq:utilization}
    \begin{split}
    \gamma_u = \frac{\mbox{Charging Time}}{\mbox{Total Time}} &= \frac{\tau\ T\ \mathbb{E}[T_{pc}-T_o]}{N\ T}\\
    &= \frac{\mathbb{E}[N_{pc}]}{N}\left(1-\frac{\mathbb{E}[T_o]}{\mathbb{E}[T_{pc}]}\right).
    \end{split}
    \end{equation}
\item \textbf{Revenue Rate:} The revenue rate $\overline{R}$ is defined as the average rate at which revenue is accrued from the park-and-charge spots, and is given by:
    \begin{equation}\label{eq:revrate}
    \overline{R} = \frac{\mbox{Total Revenue}}{\mbox{Total Time}} = \frac{\tau\ T\ \mathbb{E}[R]}{T} = \frac{\mathbb{E}[N_{pc}]\ \mathbb{E}[R]}{\mathbb{E}[T_{pc}]}.
    \end{equation}
\end{enumerate}
When $T_c$, $T_a$, $C_{\max}$ are generally distributed, and the price/penalty functions $p_c(t)$, $p_o(t)$ are general, the above quantities do not admit closed form expressions, and we defer their (partial) mathematical derivations to Appendix~\ref{appendix:computeExpr}. However, in Section~\ref{specialCase}, we investigate a simple special case where the price/penalty functions are linear, $T_c$ and $T_a$ are Exponentially distributed, and $C_{\max}$ is a constant, and derive closed form expressions for these performance measures.

\noindent\textbf{Benchmarking performance:} In order to measure the effectiveness of our penalty scheme, we benchmark it against a system with ``ideal'' human behavior, one in which the EV users, on their own, do not overstay, i.e., $T_{pc}=\min\{T_c,T_a\}$. Here, $\overline{q}=1$, and the revenue from~\eqref{eq:rev} is simply $p_{c}(T_{pc})$.

\section{Example scenario: Linear functions, Exponential distributions.}\label{specialCase}
In this section, we consider a simplified scenario for which we derive closed form expressions for the performance measures, which allows better analysis of the effectiveness of imposing simple, linear penalties for overstaying. In particular, we assume:
\begin{itemize}
\item \textit{Linearity:} Let $p_{c}(t)=\alpha_c t$ and $p_{o}(t)=\alpha_o t$, where $\alpha_c > 0$ and $\alpha_o\geq 0$ are the parameters.
\item $T_c$ and $T_{a}$ are Exponentially distributed with parameters $\mu_c$ and $\mu_a$ respectively.
\item $C_{\max}$ is not a random variable, but a constant.
\end{itemize}
Under these assumptions, the acceptance probability $q$ from~\eqref{eq:q} and its mean $\overline{q}$ from~\eqref{eq:qbar} can be evaluated to obtain $q=1-\beta e^{-\mu_a T_c}$ and $\overline{q} = 1-\frac{\beta\mu_c}{\mu_a+\mu_c}$ respectively, where
\begin{equation}\label{eq:beta}
\beta = e^{-\mu_a \frac{C_{\max}}{\alpha_o}}.
\end{equation}
\textbf{Conditional distribution of $T_c$:} While $T_c$ is Exponentially distributed by assumption, given that (with probability $q$,) an EV accepts the posted penalty and enters the parking lot (call this event $E$), the \textit{conditional} random variable $T_c|E$ need not be Exponential. ($T_{a}$ remains unchanged, since $E$ does not depend on it.) Thus, we first compute the conditional probability density function, $f_{c|E}$, using Bayes's rule, to obtain:
\begin{equation*}
\begin{split}
f_{c|E}(T_c) &= \frac{Prob(E\ |\ T_c)f_c(T_c)}{\int_{T_c=0}^{\infty}Prob(E\ |\ T_c)f_c(T_c)dT_c} = \frac{q\ f_c(T_c)}{\int_{T_c=0}^{\infty}q\ f_c(T_c)dT_c}\\
&= \frac{q\ f_c(T_c)}{\overline{q}} = \frac{1}{\overline{q}}\ \mu_c\ e^{-\mu_c T_c}\left(1-\beta e^{-\mu_a T_c}\right).
\end{split}
\end{equation*}

\subsection{Distribution and mean of \texorpdfstring{$T_{pc}$}{Tpc}.}
The complementary cumulative distribution function of $T_{pc}$, defined as $\overline{F}_{pc}(t) = Prob(T_{pc}> t)$, can be evaluated as:
\begin{equation*}
\begin{split}
\overline{F}_{pc}(t) &= Prob\left(T_a> t\right)Prob\left(T_c+p_{o}^{-1}(C_{\max})> t\ |\ E\right)\\
&= \left(1-F_a(t)\right)\left(\int_{T_c=t-\frac{C_{\max}}{\alpha_o}}^{\infty}f_{c|E}(T_c)dT_c\right)\\
&=
\begin{cases}
e^{-\mu_a t}, & 0\leq t \leq \frac{C_{\max}}{\alpha_o}\\
e^{-\mu_a t}\left(\int_{T_c=t-\frac{C_{\max}}{\alpha_o}}^{\infty}\frac{1}{\overline{q}}\ \mu_c\ e^{-\mu_c T_c}\left(1-\beta e^{-\mu_a T_c}\right)dT_c\right), & t \geq \frac{C_{\max}}{\alpha_o}
\end{cases}\\
&=
\begin{cases}
e^{-\mu_a t}, & 0\leq t \leq \frac{C_{\max}}{\alpha_o}\\
\frac{e^{-\mu_a t}}{\overline{q}}e^{-\mu_c\left(t - \frac{C_{\max}}{\alpha_o}\right)}\left(1-\frac{\mu_c}{\mu_a+\mu_c}e^{-\mu_a t}\right), & t \geq \frac{C_{\max}}{\alpha_o}
\end{cases}
\end{split}
\end{equation*}
Thus, the mean, given by $\int_{0}^{\infty}\overline{F}_{pc}(t)dt$, can be evaluated as:
\begin{equation}\label{eq:meanTpc-sp}
\begin{split}
\mathbb{E}[T_{pc}] &= \int_{0}^{C_{\max}/\alpha_o}e^{-\mu_a t}dt + \int_{C_{\max}/\alpha_o}^{\infty}\frac{e^{-\mu_a t}}{\overline{q}}e^{-\mu_c\left(t - \frac{C_{\max}}{\alpha_o}\right)}\left(1-\frac{\mu_c}{\mu_a+\mu_c}e^{-\mu_a t}\right) dt\\
&= \frac{1-\beta}{\mu_a} + \frac{e^{\mu_c C_{\max}/\alpha_o}}{\overline{q}}\left(\int_{C_{\max}/\alpha_o}^{\infty}e^{-(\mu_a+\mu_c)t}\left(1-\frac{\mu_c}{\mu_a+\mu_c}e^{-\mu_a t}\right)dt\right)\\
&= \frac{1-\beta}{\mu_a} + \frac{e^{\mu_c C_{\max}/\alpha_o}}{\overline{q}}\left(\frac{e^{-(\mu_a+\mu_c) C_{\max}/\alpha_o}}{\mu_a+\mu_c}-\frac{\mu_c e^{-(2\mu_a+\mu_c) C_{\max}/\alpha_o}}{(\mu_a+\mu_c)(2\mu_a+\mu_c)}\right)\\
&= \frac{1-\beta}{\mu_a} + \frac{\beta}{\overline{q}}\left(\frac{1}{\mu_a+\mu_c}-\frac{\beta\mu_c}{(\mu_a+\mu_c)(2\mu_a+\mu_c)}\right)\\
&= \frac{1-\beta}{\mu_a} + \frac{\beta}{\mu_a+(1-\beta)\mu_c}\left(1-\frac{\beta\mu_c}{2\mu_a+\mu_c}\right) = \frac{1-\beta}{\mu_a} + \frac{\beta}{\mu_a+(1-\beta)\mu_c}\cdot\frac{2\mu_a+(1-\beta)\mu_c}{2\mu_a+\mu_c}\\
&= \frac{1-\beta}{\mu_a} + \frac{\beta}{2\mu_a+\mu_c}\left(1+\frac{\mu_a}{\mu_a+(1-\beta)\mu_c}\right)\\
&= \frac{1}{\mu_a} - \frac{\beta}{2\mu_a+\mu_c}\left(\frac{\mu_a+\mu_c}{\mu_a}-\frac{\mu_a}{\mu_a+(1-\beta)\mu_c}\right),
\end{split}
\end{equation}
where $\beta = e^{-\mu_a C_{\max}/\alpha_o}$.

\subsection{Distribution and mean of \texorpdfstring{$T_o$}{To}.}

The complementary cumulative distribution function of $T_{o}$, defined as $\overline{F}_{o}(t) = Prob(T_{o}> t)$, can be evaluated as:
\begin{equation*}
\begin{split}
\overline{F}_o(t) &= Prob\left(T_o > t\right) = Prob\left(T_{pc}-T_{c}> t\ |\ E\right)\\
&= Prob\left(\min\left\{p_{o}^{-1}(C_{\max}),T_a-T_c\right\}> t\ |\ E\right)\\
&= Prob\left((T_a> T_c+t\ |\ E)\ \mbox{\texttt{AND}}\ (C_{\max}> \alpha_{o}t)\right)\\
&=
\begin{cases}
\int_{T_c=0}^{\infty}\int_{T_a=T_c+t}^{\infty}f_{c|E}(T_c)f_a(T_a)dT_cdT_a, & 0 \leq t < \frac{C_{\max}}{\alpha_o}\\
0, & t \geq \frac{C_{\max}}{\alpha_o}
\end{cases}\\
&=
\begin{cases}
\frac{e^{-\mu_a t}}{\overline{q}}\left(\frac{\mu_c}{\mu_a+\mu_c}-\frac{\beta\mu_c}{2\mu_a+\mu_c}\right), & 0 \leq t < \frac{C_{\max}}{\alpha_o}\\
0, & t \geq \frac{C_{\max}}{\alpha_o}
\end{cases}
\end{split}
\end{equation*}
Thus, the mean, given by $\int_{0}^{\infty}\overline{F}_{o}(t)dt$, can be evaluated as:
\begin{equation}\label{eq:meanTo-sp}
\begin{split}
\mathbb{E}[T_{o}] &= \int_{0}^{C_{\max}/\alpha_o}\frac{e^{-\mu_a t}}{\overline{q}}\left(\frac{\mu_c}{\mu_a+\mu_c}-\frac{\beta\mu_c}{2\mu_a+\mu_c}\right) dt\\
&= \frac{1-\beta}{\mu_a}\ \frac{1}{\overline{q}}\left(\frac{\mu_c}{\mu_a+\mu_c}-1+1-\frac{\beta\mu_c}{2\mu_a+\mu_c}\right) = \frac{1-\beta}{\mu_a}\ \frac{1}{\overline{q}}\left(\frac{2\mu_a+(1-\beta)\mu_c}{2\mu_a+\mu_c}-\frac{\mu_a}{\mu_a+\mu_c}\right)\\
&= \frac{1-\beta}{\mu_a}\ \frac{1}{\overline{q}}\left(\frac{\mu_a+(1-\beta)\mu_c}{2\mu_a+\mu_c}-\frac{\mu_a^2}{(\mu_a+\mu_c)(2\mu_a+\mu_c)}\right)\\
&= \frac{1-\beta}{\mu_a}\cdot\frac{\mu_a+\mu_c}{\mu_a+(1-\beta)\mu_c}\left(\frac{\mu_a+(1-\beta)\mu_c}{2\mu_a+\mu_c}-\frac{\mu_a^2}{(\mu_a+\mu_c)(2\mu_a+\mu_c)}\right)\\
&= \frac{1-\beta}{\mu_a}\cdot\frac{1}{2\mu_a+\mu_c}\left(\mu_a+\mu_c-\frac{\mu_a^2}{\mu_a+(1-\beta)\mu_c}\right)\\
&= \frac{1-\beta}{2\mu_a+\mu_c}\left(\frac{\mu_a+\mu_c}{\mu_a}-\frac{\mu_a}{\mu_a+(1-\beta)\mu_c}\right),
\end{split}
\end{equation}
where $\beta = e^{-\mu_a C_{\max}/\alpha_o}$.

\subsection{Mean revenue.}

When the pricing/penalty functions are linear, the revenue defined in~\eqref{eq:rev} is given by $R=\alpha_c (T_{pc}-T_{o}) + \alpha_o T_{o}$, and by linearity of expectations, its mean can be evaluated to obtain:
\begin{equation}\label{eq:meanrev-sp}
\begin{split}
\mathbb{E}[R] &= \alpha_c\left(\mathbb{E}[T_{pc}]-\mathbb{E}[T_{o}]\right) + \alpha_o \mathbb{E}[T_{o}]\\
&= \frac{\alpha_c}{2\mu_a+\mu_c}\left(1+\frac{\mu_a}{\mu_a+(1-\beta)\mu_c}\right) + \alpha_o \mathbb{E}[T_{o}],
\end{split}
\end{equation}
where $\beta = e^{-\mu_a C_{\max}/\alpha_o}$, and $\mathbb{E}[T_{o}]$ is given by~\eqref{eq:meanTo-sp}.

\subsection{Performance measures.}

\begin{figure}[htbp]
    \centering
    \subfigure[Utilization]{\includegraphics[width=0.26\textwidth]{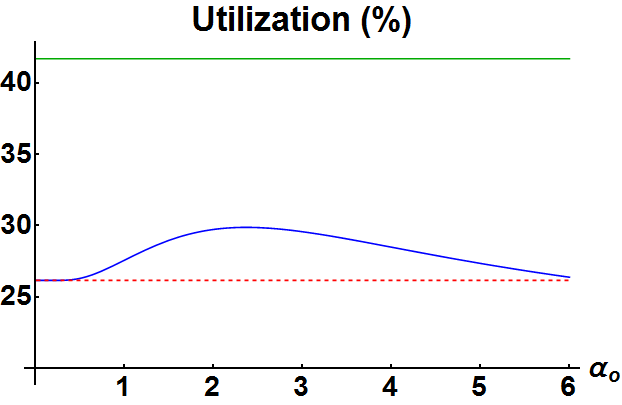}}
   \qquad
    \subfigure[Revenue Rate]{\includegraphics[width=0.26\textwidth]{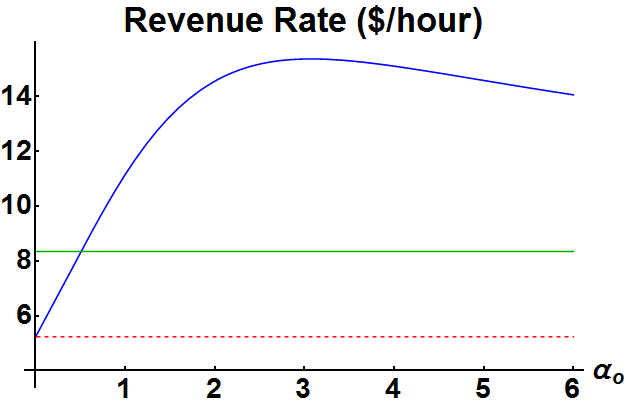}}
    \caption{Comparing the utilization and revenue rate obtained by imposing a linear penalty function with various per-hour penalty rates (blue curve), against an ideal scenario where no EV user overstays (green line). The dashed red line corresponds to the scenario where there is overstaying, but no penalty is imposed.\label{fig:exponential-performance}}
\end{figure}

By inspecting the expressions, it can be seen that $\beta$ is increasing in $\alpha_o$. Hence, with increasing penalty, $\mathbb{E}[T_{pc}]$ increases and $\mathbb{E}[T_{o}]$ decreases. The other quantities, including performance measures~\eqref{eq:throughput}-\eqref{eq:revrate}, are more complicated and their behavior is not obvious from inspection. As an example, when we set $N=10$ slots, $\lambda=8$ EVs per hour, $\mu_c=\frac{60}{45}$ per hour, $\mu_a=\frac{60}{105}$ per hour, $C_{\max}=\$4$, $\alpha_c=\$2$ per hour, we find that setting the penalty rate $\alpha_o=\$2.37$ per hour provides the maximum value of utilization ($30\%$), whereas the utilization with no penalty is about $26\%$. (The maximum possible utilization, under the ideal setting where there is no overstaying is $42\%$.) From the point of view of revenue, setting the penalty rate $\alpha_o=\$3.07$ per hour provides the maximum revenue rate of $\$15.36$ per hour, but the decreased utilization at this penalty rate is only slightly less, at $29.5\%$. (The ideal revenue when there is no overstaying is merely $\$8.34$ per hour.) Thus, the revenue maximizing penalty rate provides increased utilization (by $4\%$), as well as significantly increased revenue (by $85\%$). (See Figure~\ref{fig:exponential-performance}.)

\section{Dynamic environment.}

In this section, we consider a real world setting where no prior information about the distributions $F_c$, $F_a$, $F_{\max}$ is known, and thus, the expressions for the performance measures are also unknown to the framework. We assume that these distributions are unknown, but fixed for the day; hence, the utilization and revenue for the day depend only on $\alpha_o$.

In our model, on each day, a penalty rate $\alpha_o$ is declared at the entrance to the parking area. Arriving customers behave according to their $T_c$, $T_a$, $C_{max}$ values (which are unknown to the system, but known to the customers), and the posted penalty rate $\alpha_o$; their behavior is as described in Section~\ref{ssec:user-behavior}. The system observes the customer behavior only after they enter the parking area. Thus, the performance measures (such as revenue rate, utilization, etc.) corresponding to penalty rate $\alpha_o$ are only observed at the end of the day. The task is to find an optimal penalty rate $\alpha_o^*$ for each day, such that the total revenue over all days is maximized.\footnote{Variants of the technique presented here can be used for other performance measures such as utilization.}

This problem is an example of sequential decision making in an unknown and dynamic environment, where the system seeks to optimize the average revenue over all days, while continuously gathering more information about the revenue obtained by using different penalty rates $\alpha_o$. This leads to a trade-off between exploration (imposing each penalty rate $\alpha_o$ sufficiently often to obtain better estimates of the corresponding revenue accumulated) and exploitation (frequently imposing the optimal penalty for which the observed revenue is maximized). Such problems naturally fall into the category of stochastic multi-armed bandit (MAB) problems~\cite{bubeck2012regret}. Each penalty rate $\alpha_o$ is considered as an arm, and the daily revenue obtained by imposing that penalty rate is analogous to the reward obtained by pulling the corresponding arm. The goal in the MAB setting is to determine the arm to be pulled each time in order to maximize the total reward obtained.

\subsection{Learning algorithm UCB-PC.}
Let $A$ denote a finite, ordered set of penalty rates, and let $\mathcal{R}_i$ denote the expected daily revenue corresponding to penalty rate $A[i]$, which is unknown. The (unknown) optimal penalty rate is thus given by $A[i^*]$, where $i^* = \mathop{\arg\max}_i\ \mathcal{R}_i$. In order to estimate $\{\mathcal{R}_i\}$, on each day, we impose a penalty rate and observe the daily revenue, which is the sum of the payments made by the customers who choose to enter the parking area that day.\footnote{We observe, for each customer, whether they choose to enter the parking area or not, and if they do, their charging and overstaying times, as well as their final payment.} In doing so, we keep track of the observed average daily revenues $\{\hat{\mathcal{R}}_i\}$, which are then used to gradually learn the optimal penalty rate $A[i^*]$ for the parking area. Algorithm~\ref{algo:UCB-PC} is based on the techniques used in the $\mathtt{UCB1}$ algorithm~\cite{auer2002finite}, which is a well known tool for solving stochastic multi-armed bandit problems. However, $\mathtt{UCB1}$ assumes that the system is allowed to run only for a finite number of trials $T$ (``finite horizon multi-armed bandit''), whereas we operate under the assumption that the system runs for infinite time, and must therefore tackle the challenges involved in adapting $\mathtt{UCB1}$ to the infinite horizon setting.\footnote{While $\mathtt{UCB1}$ was our choice to illustrate the technique, we believe that our framework can also accommodate modified versions of other stochastic multi-armed bandit algorithms (such as $EXP3$).}

\begin{algorithm}
\caption{UCB-PC \label{algo:UCB-PC}}
\begin{algorithmic}[1]
\State \textbf{Input:} A finite set $A$ of penalty rates
\State $R_i$ stores the total observed revenue for $A[i]$
\State $K_i$ stores the number of days $A[i]$ is imposed
\For{$t \gets 1$ \textbf{to} $|A|$}
    \State Choose penalty rate $A[t]$ on $t^{th}$ day
    \State Observe revenue earned $r$
    \State Set $R_t\leftarrow r$
    \State Set $K_t\leftarrow 1$
\EndFor
\State Find $i^* = \mathop{\arg\max}_i\ \left({R_i} + \sqrt{2\ln{|A|}}\right)$
\For{$t \gets |A|+1, |A|+2, \ldots$ }
    \State Choose penalty rate $A[i^*]$ on $t^{th}$ day
    \State Observe revenue earned $r$
    \State Update $R_{i^*}\leftarrow R_{i^*}+r$
    \State Update $K_{i^*}\leftarrow K_{i^*}+1$
    \State Update $\hat{\mathcal{R}}_{i^*}\leftarrow \frac{R_{i^*}}{K_{i^*}}$
    \State Find $i^* = \mathop{\arg\max}_i\ \left(\hat{\mathcal{R}}_{i} + \sqrt{\frac{2\ln{t}}{K_i}}\right)$ for $(t+1)^{th}$ day
\EndFor
\end{algorithmic}
\end{algorithm}

\subsection{Performance analysis.}\label{sec:performance}

In this section, we compare the performance of our learning algorithm with an optimal algorithm that knows the expected revenues $\{\mathcal{R}_i\}$ beforehand, and can therefore choose the optimal penalty $A[i^*]$ every day. The loss incurred by the learning algorithm over $K$ days is termed as the \textit{regret} $\mathcal{L}_K$, and can be written as
\begin{equation}\label{eq:regret}
\mathcal{L}_K = K\cdot\mathcal{R}_{i^*} - \left(\displaystyle\sum_{i=1}^{|A|}\ \mathbb{E}[K_i(K)] \mathcal{R}_i\right)=\displaystyle\sum_{i=1}^{|A|}\ \mathbb{E}[K_i(K)] ( \mathcal{R}_{i^*} - \mathcal{R}_i),
\end{equation}
where $\mathbb{E}[K_i(K)]$ denotes the expected number of days (out of $K$) that penalty rate $A[i]$ is chosen by our learning algorithm. In the literature, the performance of multi-armed bandit algorithms is measured by obtaining an upper bound on this regret. In~\cite{auer2002finite}, the authors show that the upper bound on the regret using $\mathtt{UCB1}$ is sublinear within a finite, fixed horizon $T$; in particular, they show that the regret is bounded above by a term that is $\tilde{O}(\log{T})$. Next, in Theorem~\ref{theorem:regret-bound}, we show that the regret for $\mathtt{UCB}$-$\mathtt{PC}$ is upper bounded by $\tilde{O}(\log{K})$ after the algorithm runs for $K$ days, for all $K$.

\begin{theorem}\label{theorem:regret-bound}
The expected regret of $\mathtt{UCB}$-$\mathtt{PC}$, after the algorithm runs for $K$ days, is $\displaystyle\sum_{\substack{i=1\\ i\neq i^*}}^{|A|}\left(\left\lceil\frac{8\ln{K}}{(\mathcal{R}_{i^*}-\mathcal{R}_i)^2}\right\rceil + 1 + \frac{\pi^2}{3}\right)(\mathcal{R}_{i^*}-\mathcal{R}_i)$.
\end{theorem}

\proof{Proof Sketch.} The proof follows from~\cite{auer2002finite}, where the upper bound for regret is obtained by bounding the expected number of times an arm is pulled. In our case, the upper bound on $\mathbb{E}[K_i(K)]$ for each $i\neq i^*$ is given by:
\begin{equation}\label{eq:EKi}
\mathbb{E}[K_i(K)]\leq \tau + \displaystyle \sum_{t=\tau}^{K-1} \mathbb{P}\left(A[i]\mbox{ is imposed on day }(t+1)\wedge\ K_i(t) \geq \tau \right),
\end{equation}
for any finite positive integer $\tau$. Next, in Lemma~\ref{lemma2}, we show that, for any suboptimal penalty index $i\neq i^*$, the term $\mathbb{P}\left(A[i]\mbox{ is imposed on day }(t+1)\wedge\ K_i(t) \geq \tau \right)$ is decreasing in $t$ after sufficient exploration.

\begin{lemma}\label{lemma2}
For any $i\neq i^*$, if penalty rate $A[i]$ is chosen for at least $\left\lceil\frac{8\ln{t}}{(\mathcal{R}_{i^*}-\mathcal{R}_i)^2}\right\rceil$ days among the first $t$ days, then
\begin{equation*}
\mathbb{P}\left(A[i]\mbox{ is imposed on day }(t+1)\ \wedge\ K_i(t) \geq \tau \right) \leq \frac{2}{t^2}.
\end{equation*}
\end{lemma}

\proof{Proof.} We begin by writing,
\begin{equation}\label{eq:lem1}
\begin{split}
&\mathbb{P}\left(A[i]\mbox{ is imposed on day }(t+1)\ \wedge\ K_i(t) \geq \tau \right)\\
&\qquad=\displaystyle \sum_{s'=1}^{t} \sum_{s=\tau+1}^{t} \mathbb{P}\left(A[i]\mbox{ is imposed on day }(t+1)\wedge \ K_i(t) = s\ \wedge \ K_{i^*}(t)=s'\right).
\end{split}
\end{equation}
After running $\mathtt{UCB}$-$\mathtt{PC}$ for $t$ days, a suboptimal penalty index $i \neq i^*$ is chosen only when $\hat{\mathcal{R}}_{i^*} + \sqrt{\frac{2\ln{t}}{s'}} \leq \hat{\mathcal{R}}_{i} + \sqrt{\frac{2\ln{t}}{s}}$. Let this event be denoted by $W$. Also, let $U$ denote the event $\left(\hat{\mathcal{R}}_{i^*} \leq \mathcal{R}_{i^*} - \sqrt{\frac{2\ln{t}}{s'}}\right)$, and $V$ denote the event $\left(\hat{\mathcal{R}}_{i} \geq \mathcal{R}_{i} + \sqrt{\frac{2\ln{t}}{s}}\right)$. Now, the probability of choosing penalty rate $A[i]$ on the $(t+1)^{th}$ day can be written as:
\begin{equation}\label{eq:PW}
\mathbb{P}\left(W\right)\leq \mathbb{P}(W|U)\mathbb{P}(U)+\mathbb{P}(W|V)\mathbb{P}(V)+\mathbb{P}(W|\bar{U}\wedge\bar{V})\mathbb{P}(\bar{U}\wedge\bar{V}).
\end{equation}
Using the Chernoff-Hoeffding's inequality, the following can be shown:
\begin{equation}\label{eq:PU}
\mathbb{P}(U) = \mathbb{P}\left(\hat{\mathcal{R}}_{i^*} \leq \mathcal{R}_{i^*} - \sqrt{\frac{2\ln{t}}{s'}}\right) \leq\ e^{-2\left(\sqrt{\frac{2\ln{t}}{s'}}\right)^2s'} = \frac{1}{t^4}.
\end{equation}
\begin{equation}\label{eq:PV}
\mathbb{P}(V)= \mathbb{P}\left(\hat{\mathcal{R}}_{i} \geq \mathcal{R}_{i} + \sqrt{\frac{2\ln{t}}{s}}\right)\leq\frac{1}{t^4}.
\end{equation}
Next, we show that
\begin{equation}\label{eq:W}
\mathcal{P}(W|\bar{U}\wedge\bar{V})=0 \mbox{ if } s \geq \frac{8\ln{t}}{(\mathcal{R}_{i^*}-\mathcal{R}_i)^2}.
\end{equation}
First, we observe that the events $\bar{U}$ and $\bar{V}$ can be equivalently written as:
\begin{equation}\label{eq:UV}
\begin{split}
\bar{U}\Leftrightarrow&\hat{\mathcal{R}}_{i^*} > \mathcal{R}_{i^*} - \sqrt{\frac{2\ln{t}}{s'}} \Leftrightarrow  \mathcal{R}_{i^*}<\hat{\mathcal{R}}_{i^*} + \sqrt{\frac{2\ln{t}}{s'}}\\
\bar{V}\Leftrightarrow&\hat{\mathcal{R}}_{i} < \mathcal{R}_{i} + \sqrt{\frac{2\ln{t}}{s}} \Leftrightarrow \hat{\mathcal{R}}_{i} + \sqrt{\frac{2\ln{t}}{s}}<\mathcal{R}_{i} + 2\sqrt{\frac{2\ln{t}}{s}}\\
\end{split}
\end{equation}
Then, we show that if~\eqref{eq:UV} holds and $s \geq \frac{8\ln{t}}{(\mathcal{R}_{i^*}-\mathcal{R}_i)^2}$, then the event $W$ is a contradiction:
\begin{equation*}
\begin{split}
W\Leftrightarrow&\hat{\mathcal{R}}_{i^*} + \sqrt{\frac{2\ln{t}}{s'}}  \leq  \hat{\mathcal{R}}_{i} + \sqrt{\frac{2\ln{t}}{s}}\\
\Rightarrow& \mathcal{R}_{i^*} < \mathcal{R}_i + 2\sqrt{\frac{2\ln{t}}{s}}\\
\Rightarrow& \mathcal{R}_{i^*} < \mathcal{R}_i + \sqrt{(\mathcal{R}_{i^*}-\mathcal{R}_i)^2}\\
\Rightarrow& \mathcal{R}_{i^*} < \mathcal{R}_{i^*},
\end{split}
\end{equation*}
which is a contradiction. Finally, substituting the values obtained by~\eqref{eq:PU}-\eqref{eq:W} in~\eqref{eq:PW}, we get $\mathbb{P}(W) \leq \frac{2}{t^4}$, which in turn, can be substituted in~\eqref{eq:lem1} to obtain
\begin{equation*}
\mathbb{P}(A[i] \mbox{ imposed on } (t+1)^{th} day)\leq\displaystyle \sum_{s'=1}^{t} \sum_{s=\tau+1}^{t} \frac{2}{t^4} \leq\frac{2}{t^2}.
\end{equation*}
This completes the proof.
\eProof

Using Lemma~\ref{lemma2} and substituting $\tau=\left\lceil\frac{8\ln{K}}{(\mathcal{R}_{i^*}-\mathcal{R}_i)^2}\right\rceil$ in~\eqref{eq:EKi}, we obtain the required upper bound on regret that is specified by Theorem~\ref{theorem:regret-bound}. It can be shown that $\mathcal{L}_K = \tilde{O}(\ln{K})$ for a fixed set of penalty rates. Thus, the average daily regret, given by $\frac{\mathcal{L}_K}{K}$, vanishes as $K\rightarrow\infty$.

\section{Experimental results.}

\begin{figure}[htbp]
    \centering
    \subfigure[Histogram of parking duration data]{\label{fig:hist_Ta}\includegraphics[width=0.35\textwidth]{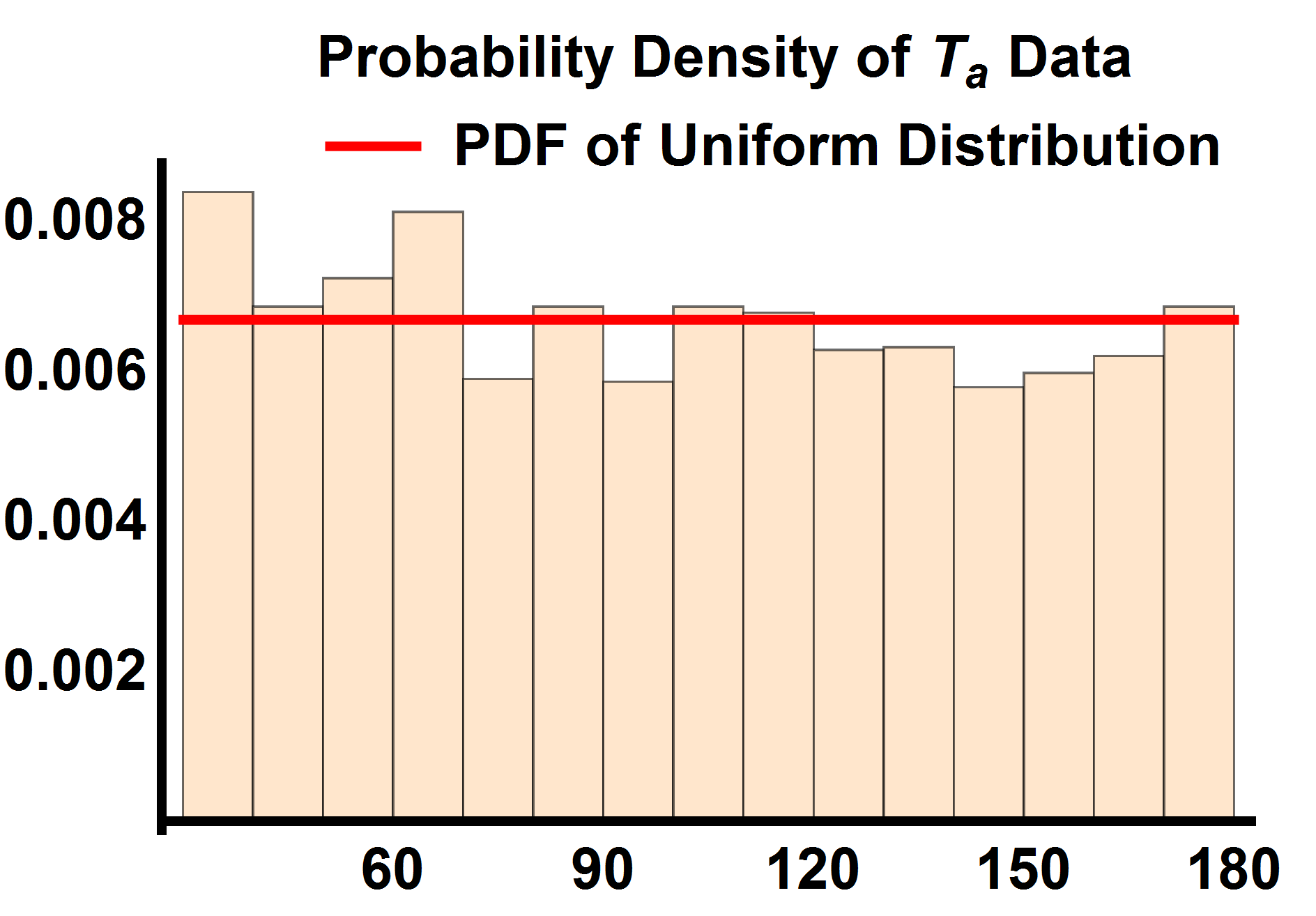}}
   \qquad
    \subfigure[Histogram of charging duration data]{\label{fig:hist_Tc}\includegraphics[width=0.35\textwidth]{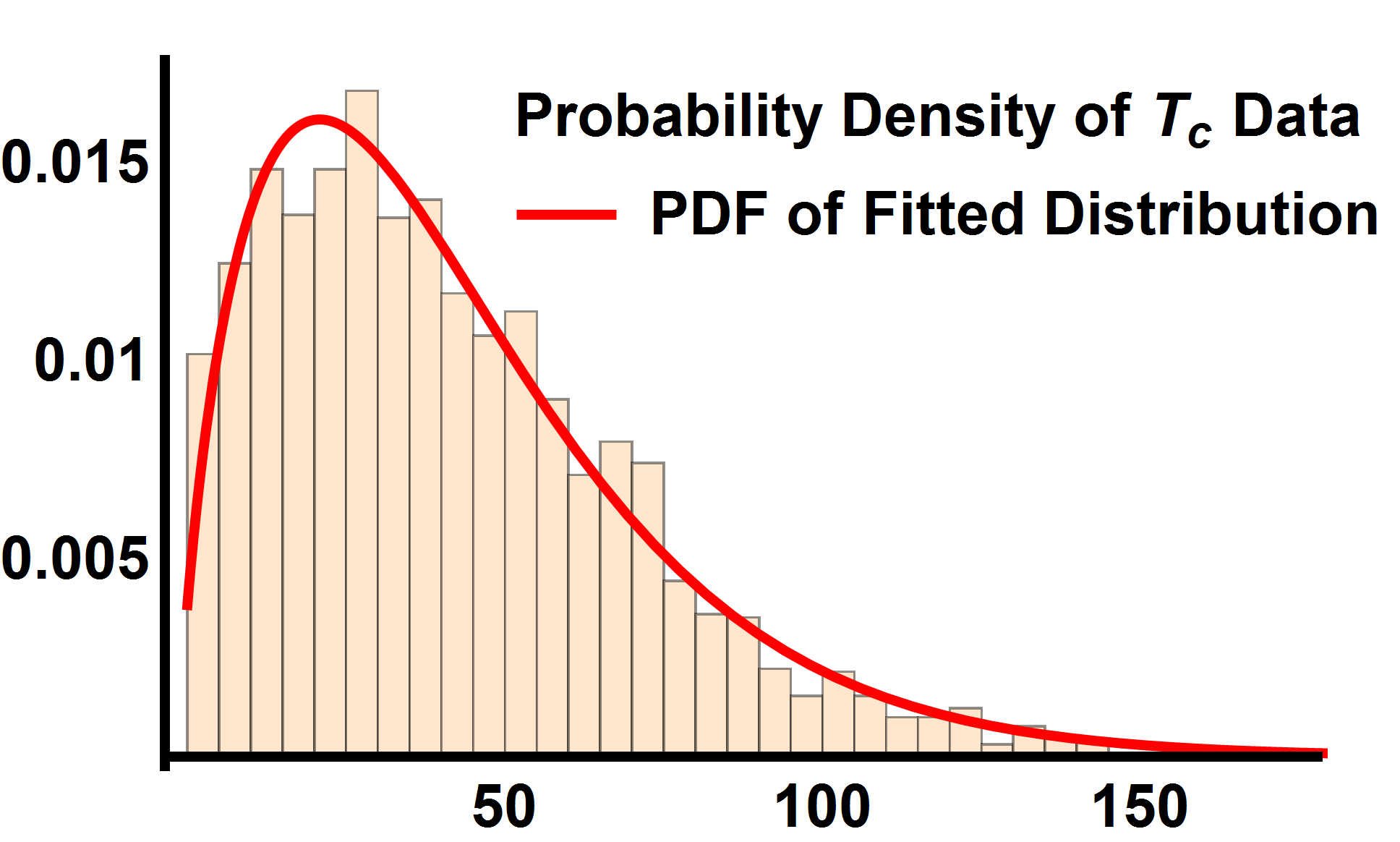}}
    \caption{Histograms of real-world data for London, restricted to ``standard'' chargers and parking duration between $30$ and $180$ minutes.}
    \vspace{-0.1in}
\end{figure}

For simulating the behavior of electric vehicles we study real-world data for the city of London, obtained from~\cite{Plugged-in-places}, which consists of 9961 charging events, including usage data of charge points and the duration of stay for EVs. We find it difficult to fit a single parametric model to these data without any restrictions.\footnote{Mixture models can be quite effective here, because they reveal the hidden heterogeneity that arises from a latent categorical variable such as charging point type, parking duration (very short, standard, long), time of arrival (peak hours and non-peak hours), etc.} Therefore, we only look at data entries that correspond to the ``standard'' charging point type and a specific duration of parking time (between $30$ minutes and $180$ minutes), in order to fit them in a single parametric model. We assume that the charging duration data corresponds to $T_c$ and the parking duration data corresponds to $T_a$, the duration for which customers would park their car in the absence of any penalties. The histogram for the restricted parking duration data and charging duration data are shown in Figures~\ref{fig:hist_Ta} and~\ref{fig:hist_Tc}.

Using Mathematica, we obtain the distributions that best fit these histograms. This corresponds to a uniform distribution between $30$ minutes and $180$ minutes for $T_a$, and a generalized Gamma distribution with parameters $\mu=-1.35188$, scale parameter $33.7831$, and shape parameters $1.44212$ and $1.19403$, for $T_c$.

We assume that the penalty threshold $C_{max}$ is a discrete random variable that takes values $\$4$, $\$8$, $\$10$, $\$20$ with probabilities $0.4$, $0.3$, $0.2$, $0.1$ respectively. We also assume a linear pricing function during charging, with rate $\alpha_c= \$2$ per hour. We simulate vehicles arriving to a parking area with $10$ charging slots for a $6$ hour time period, assuming that the arrivals follow a Poisson distribution with rate $10$ per hour. We also assume that when all the slots of a parking lot are occupied, arriving users do not wait for a slot to be free and immediately leave the parking area. We perform two sets of experiments, discussed next.

\subsection{Experiment 1.}

\begin{figure}[htbp]
    \centering
    \subfigure[Utilization]{\includegraphics[width=0.35\textwidth]{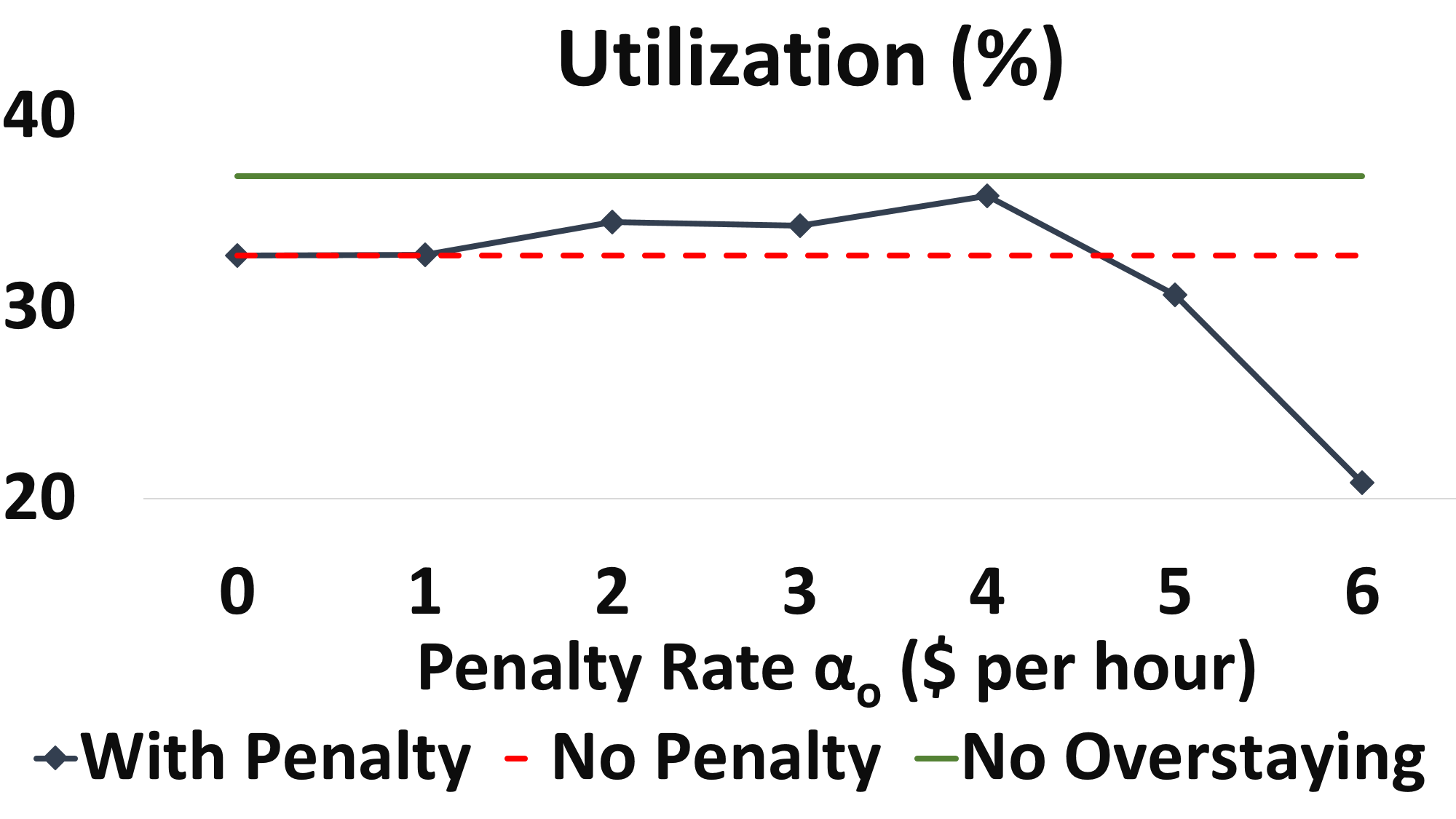}}
   \qquad
    \subfigure[Revenue]{\includegraphics[width=0.35\textwidth]{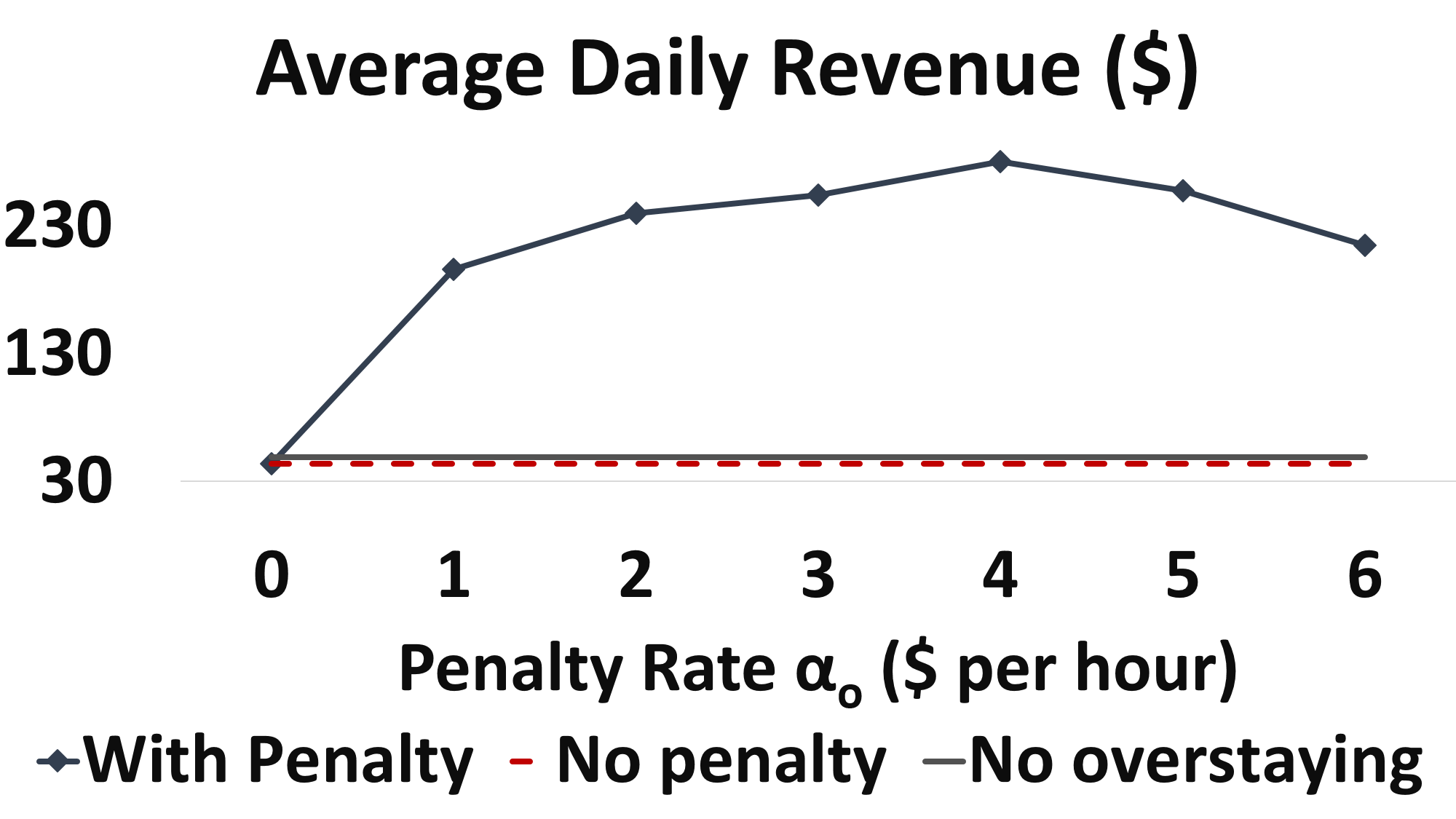}}
    \caption{Comparing the utilization and total revenue obtained by imposing a linear penalty function with various per-hour rates, against an ideal scenario where no EV user overstays.\label{fig:obs1}}
    \vspace{-0.1in}
\end{figure}

We simulate the parking area for a $6$ hour time period per day for $100$ days. Then, we observe the utilization and revenue obtained for each day by employing a linear penalty function $p_o(t) = \alpha_o*t$ with $\alpha_o\in\{0,1,2,3,4,5,6\}$. Note that $\alpha_o=0$ corresponds to the ``no-penalty'' scenario. We compare these results against an ideal benchmark where we assume that EV users do not overstay at all. Figure~\ref{fig:obs1} shows the results averaged over the $100$ days.

\noindent\textbf{Utilization:} The utilization of the parking area increases (dark blue line) after imposing a small penalty as compared to when no penalty (red dotted line) is charged. However, when a high penalty is imposed, most of the EV users decide not to enter the parking area, resulting in a steep drop in utilization. With a penalty rate $\$4$ per hour, the utilization is maximum, and also very close to the ideal utilization when there is no overstaying by the EV users (green line). Thus, imposing just a small penalty helps improve the utilization of charging spots in the parking area significantly.\\

\noindent\textbf{Revenue:} The daily revenue from the parking area quickly increases even after imposing a small penalty for overstaying. The daily revenue obtained in the ideal situation of no overstaying is more than when no penalty is imposed (albeit only slightly so), since more customers are served in the former situation than the latter. As the penalty rate increases, the daily revenue follows the same pattern as the utilization, and for the same reasons. It should be noted that the penalty rate $\$4$ per hour also maximizes the daily revenue.

\subsection{Experiment 2.}

\begin{figure}[htbp]
    \centering
    \subfigure[Daily Revenue]{\includegraphics[width=0.4\textwidth]{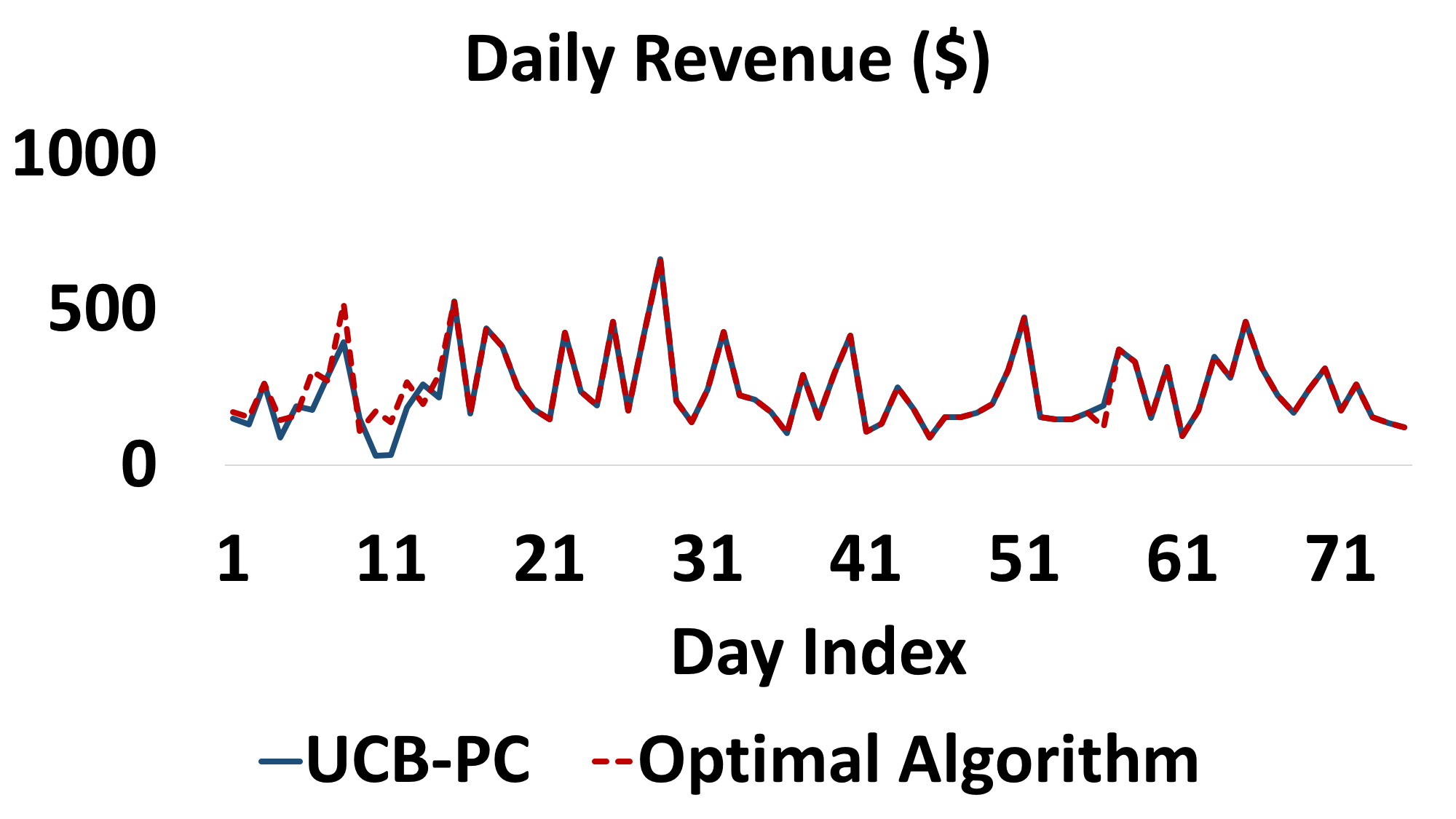}}
   \qquad
    \subfigure[Average Difference in Revenue]{\includegraphics[width=0.4\textwidth]{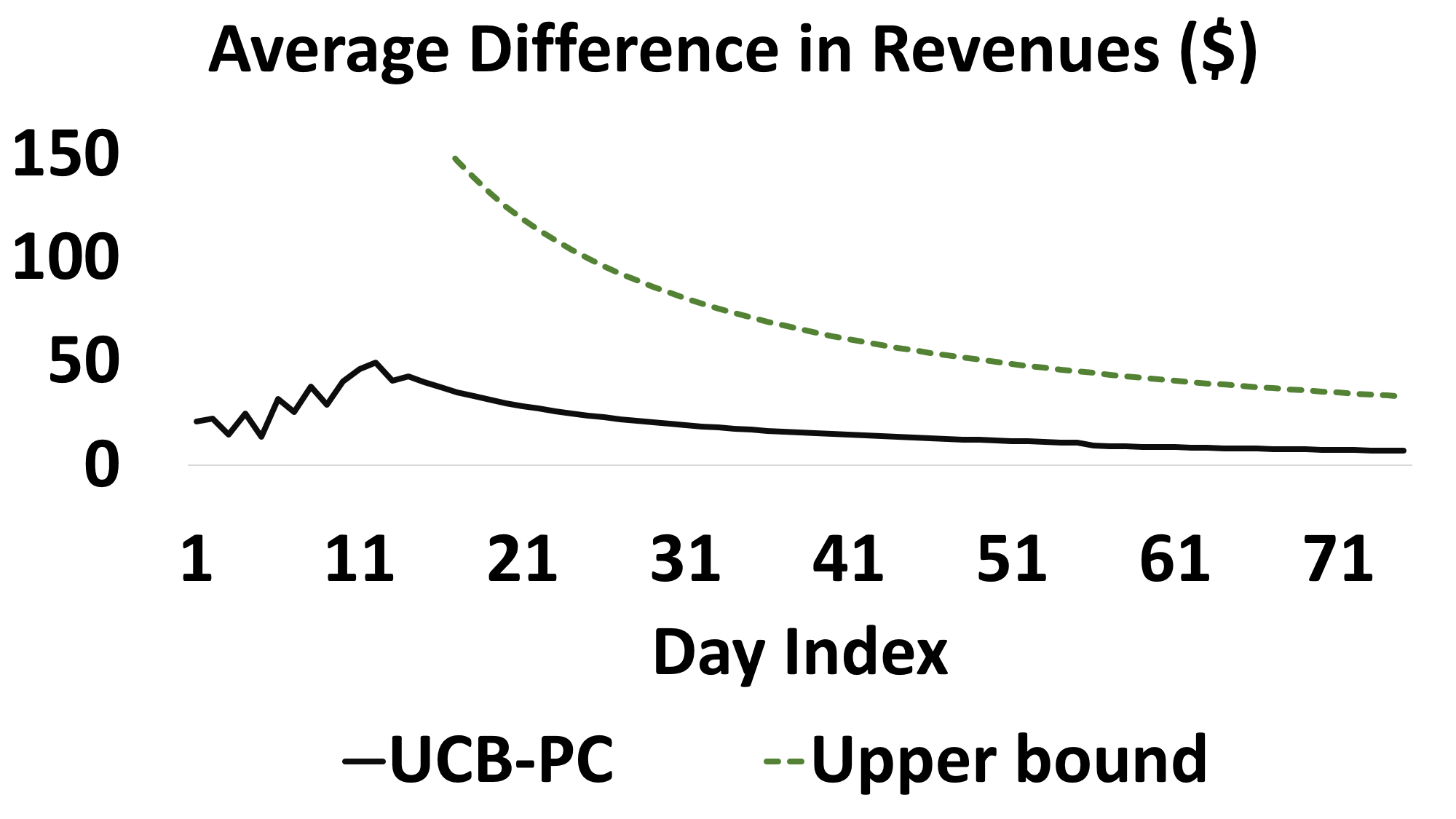}}
    \caption{Comparing the performance of learning algorithm with respect to the optimal algorithm\label{fig:obs2}}
\end{figure}

We simulate the parking area for two different settings: (a)~the optimal penalty rate $\alpha_o^*$ is chosen on all days, where $\alpha_o^*$ is calculated by approximately) maximizing the total expected revenue with complete knowledge of all the underlying distributions that are used, and (b)~our learning algorithm $\mathtt{UCB}$-$\mathtt{PC}$ determines the best $\alpha_o$ for each day, based on the observed parameters for all previous days. We compare the convergence of the average revenue obtained by our learning method to the average revenue obtained by the (approximately) optimal method. Figure~\ref{fig:obs2} shows the results.

We observe that the penalty rates chosen by our learning algorithm $\mathtt{UCB}$-$\mathtt{PC}$ are almost equivalent to that of the optimal algorithm after $15$ days in terms of the daily revenue obtained. We also show the difference between the total revenue by the optimal algorithm and $\mathtt{UCB}$-$\mathtt{PC}$ averaged over the number of days, along with the theoretical upper bound given in Theorem~\ref{theorem:regret-bound}. As expected, it can be seen that the average difference, that is, $\frac{\mathcal{L}_k}{k}$, decreases with increase in the number of days after sufficient exploration (15 days).

\section{Concluding remarks.}

In this paper, we undertake a formal study, for the first time, of the problem of overstaying EVs in park-and-charge spots. We establish a novel framework in which we bring together an interdisciplinary mix of models and techniques: probabilistic user behaviour, queueing dynamics, online learning. This framework can be extended to accommodate different user behaviour models, queueing dynamics, and other learning techniques. One can imagine ``instantiating'' it for multiple parking areas in a city with a model for the population of EV users and study the interaction at a higher level, e.g., competition between parking lots. When viewed as a comprehensive model for park-and-charge, our framework could be a useful tool for future research.

\bibliographystyle{ormsv080}
\bibliography{submission-parkcharge}

\newpage

\begin{APPENDICES}

\section{Computing \texorpdfstring{$\mathbb{E}[T_{pc}]$}{E[Tpc]}, \texorpdfstring{$\mathbb{E}[T_{o}]$}{E[To]}, \texorpdfstring{$\mathbb{E}[R]$}{E[R]}.}\label{appendix:computeExpr}

All three quantities involve the random variables $T_{c}$, $T_{a}$, and $C_{\max}$. For convenience, we assume that they are continuous and admit probability density functions $f_c$, $f_a$, and $f_{\max}$ respectively. However, a similar approach would apply if any of the distributions were discrete or mixed-type.

\subsection{Conditional joint distribution of \texorpdfstring{$T_c$}{Tc} and \texorpdfstring{$C_{\max}$}{Cmax}.}\label{ssec:condTc}
While $T_c$ and $C_{\max}$ are independent random variables, given that an EV accepts the posted penalty and enters the parking lot (call this event $E$), the \textit{conditional} random variables $T_c|E$ and $C_{max}|E$ could be correlated and have different distributions, since $Prob(E)=q(T_c,C_{\max})$, given by~\eqref{eq:q}. ($T_{a}$ remains independent and unchanged, since $E$ does not depend on it.) Since $T_{pc}$, $T_o$, and $R$ are all properties of the EVs that have accepted the posted penalty, it becomes necessary to first compute the joint probability density function of $(T_c,C_{\max})|E$, using Bayes's rule:
\begin{equation}\label{eq:condTcCmax}
\begin{split}
f_{c\max|E}(T_c,C_{\max}) &= \frac{Prob\left(E\ |\ T_c, C_{\max}\right)f_{c}(T_c)f_{\max}(C_{\max})}{\int\int Prob\left(E\ |\ T_c, C_{\max}\right)dF_{c}dF_{\max}}\\
&= \frac{q(T_c,C_{\max})f_{c}(T_c)f_{\max}(C_{\max})}{\int\int q(T_c,C_{\max})dF_{c}dF_{\max}} = \frac{q(T_c,C_{\max})}{\overline{q}}f_{c}(T_c)f_{\max}(C_{\max}),
\end{split}
\end{equation}
where $\overline{q}$ is given by~\eqref{eq:qbar}.

\subsection{Distribution and mean of \texorpdfstring{$T_{pc}$}{Tpc}.}

The complementary cumulative distribution function of $T_{pc}$, defined as $\overline{F}_{pc}(t) = Prob(T_{pc}> t)$, can be evaluated as:
\begin{equation}\label{eq:distTpc}
\begin{split}
\overline{F}_{pc}(t) &= Prob\left(T_a> t\right)Prob\left(T_c+p_{o}^{-1}(C_{\max})> t\ |\ E\right)\\
&= \left(1-F_a(t)\right)\Big(\int_{T_c=t}^{\infty}\int_{C_{\max}=0}^{\infty}f_{c\max|E}(T_c,C_{\max})dT_c dC_{\max}\\
&\quad +\int_{T_c=0}^{t}\int_{C_{\max}=p_{o}(t-T_c)}^{\infty}f_{c\max|E}(T_c,C_{\max})dT_c dC_{\max}\Big)\\
&= \left(\frac{1-F_a(t)}{\overline{q}}\right)\Big(\int_{T_c=t}^{\infty}\int_{C_{\max}=0}^{\infty}q(T_c,C_{\max})dF_c dF_{\max}\\
&\qquad\qquad\qquad\qquad\qquad\qquad+\int_{T_c=0}^{t}\int_{C_{\max}=p_{o}(t-T_c)}^{\infty}q(T_c,C_{\max})dF_c dF_{\max}\Big).\\
\end{split}
\end{equation}
The mean is then given by $\mathbb{E}[T_{pc}]=\int_{0}^{\infty}\overline{F}_{pc}(t)dt$.

\subsection{Distribution and mean of \texorpdfstring{$T_{o}$}{To}.}

The complementary cumulative distribution function of $T_{o}$, defined as $\overline{F}_{o}(t) = Prob(T_{o}> t)$, can be evaluated, for $t\geq 0$, as:
\begin{equation*}
\begin{split}
\overline{F}_o(t) &= Prob\left(T_{pc}-T_{c}> t\ |\ E\right)\\
&= Prob\left(\min\left\{p_{o}^{-1}(C_{\max}),T_a-T_c\right\}> t\ |\ E\right)\\
&= Prob\left((T_a> T_c+t\ |\ E)\ \mbox{\texttt{AND}}\ (C_{\max}> p_{o}(t))\right)\\
&= \int_{T_c=0}^{\infty}\int_{C_{\max}=p_{o}(t)}^{\infty}\int_{T_a=T_c+t}^{\infty}f_{c\max|E}(T_c,C_{\max})dT_cdC_{\max}dF_a\\
&= \frac{1}{\overline{q}}\int_{T_c=0}^{\infty}\int_{C_{\max}=p_{o}(t)}^{\infty}\left(1-F_a(T_c+t)\right)q\left(T_c,C_{\max}\right)dF_cdF_{\max}.
\end{split}
\end{equation*}
The mean is then given by $\mathbb{E}[T_o]=\int_{0}^{\infty}\overline{F}_o(t)dt$.

\subsection{Mean revenue.}

The formula for revenue in~\eqref{eq:rev} can be alternatively written as:
\begin{equation*}
R = \begin{cases}
p_{c}(T_a), & 0\leq T_a\leq T_c\\
p_{c}(T_c)+p_{o}(T_a-T_c), & T_c\leq T_a\leq T_c+p_o^{-1}(C_{\max})\\
p_{c}(T_c)+C_{\max}, & T_a\geq T_c+p_o^{-1}(C_{\max})
\end{cases}
\end{equation*}
Thus, the average revenue can be computed as follows:
\begin{equation*}
\begin{split}
\mathbb{E}[R] &= \int_{T_c=0}^{\infty}\int_{C_{\max}=0}^{\infty}\int_{T_a=0}^{T_c}p_{c}(T_a)f_{c\max|E}(T_c,C_{\max})dT_c dC_{\max}dF_a\\
&+ \int_{T_c=0}^{\infty}\int_{C_{\max}=0}^{\infty}\int_{T_a=T_c}^{T_c+p_{o}^{-1}(C_{\max})}(p_{c}(T_c)+p_{o}(T_a-T_c))f_{c\max|E}(T_c,C_{\max})dT_c dC_{\max}dF_a\\
&+ \int_{T_c=0}^{\infty}\int_{C_{\max}=0}^{\infty}\int_{T_a=T_c+p_{o}^{-1}(C_{\max})}^{\infty}(p_{c}(T_c)+C_{\max})f_{c\max|E}(T_c,C_{\max})dT_c dC_{\max}dF_a.
\end{split}
\end{equation*}
Rearranging the terms, we get:
\begin{equation}\label{eq:meanrev}
\begin{split}
\mathbb{E}[R] &=\frac{1}{\overline{q}}\int_{T_c=0}^{\infty}\int_{C_{\max}=0}^{\infty}\int_{T_a=0}^{T_c}p_{c}(T_a)q(T_c,C_{\max})dF_c dF_{\max}dF_a\\
&+\frac{1}{\overline{q}}\int_{T_c=0}^{\infty}\int_{C_{\max}=0}^{\infty}\int_{T_a=T_c}^{\infty}p_{c}(T_c)q(T_c,C_{\max})dF_c dF_{\max}dF_a\\
&+\frac{1}{\overline{q}}\int_{T_c=0}^{\infty}\int_{C_{\max}=0}^{\infty}\int_{T_a=T_c}^{T_c+p_{o}^{-1}(C_{\max})}p_{o}(T_a-T_c)q(T_c,C_{\max})dF_c dF_{\max}dF_a\\
&+\frac{1}{\overline{q}}\int_{T_c=0}^{\infty}\int_{C_{\max}=0}^{\infty}\int_{T_a=T_c+p_{o}^{-1}(C_{\max})}^{\infty}C_{\max}q(T_c,C_{\max})dF_c dF_{\max}dF_a.\\
\end{split}
\end{equation}

\end{APPENDICES}

%%%%%%%%%%%%%%%%%
\end{document}